\documentclass{article}

\usepackage{microtype}
\usepackage{graphicx}
\usepackage{subfigure}
\usepackage{booktabs} 

\usepackage{hyperref}
\usepackage{bbm}
\usepackage{float}
\restylefloat{table}

\usepackage[accepted]{icml2023}

\usepackage{amsmath}
\usepackage{amssymb}
\usepackage{mathtools}
\usepackage{amsthm}

\usepackage[capitalize,noabbrev]{cleveref}

\theoremstyle{plain}

\theoremstyle{definition}

\theoremstyle{remark}

\usepackage[textsize=tiny]{todonotes}

\icmltitlerunning{Multi-Class Deep SVDD
}

\begin{document}

\twocolumn[
\icmltitle{Multi-Class Deep SVDD: Anomaly Detection Approach in Astronomy with Distinct Inlier Categories}



\icmlsetsymbol{equal}{*}

\begin{icmlauthorlist}
\icmlauthor{M. Pérez-Carrasco}{uds,udec,mas}
\icmlauthor{G. Cabrera-Vives}{udec,uds,mas}
\icmlauthor{L. Hernández-García}{uv,mas}
\icmlauthor{F. F\"orster}{mas,cmm,dai}
\icmlauthor{P. Sánchez-Sáez}{eso,mas}
\icmlauthor{A. Muñoz Arancibia}{mas,cmm}
\icmlauthor{N. Astorga}{die,mas}
\icmlauthor{F. E. Bauer}{mas,astropuc,astroing}
\icmlauthor{A. Bayo}{uv,eso}
\icmlauthor{M. Cádiz-Leyton}{udec,uds}
\icmlauthor{M. Catelan}{mas,astropuc,astroing}
\icmlauthor{P. A. Estévez}{mas,die}
\end{icmlauthorlist}

\icmlaffiliation{uds}{Data Science Unit, Universidad de Concepción, Concepción, Chile}

\icmlaffiliation{udec}{Department of Computer Science, Universidad de Concepción, Concepción, Chile}

\icmlaffiliation{mas}{Millennium Institute of Astrophysics (MAS), Santiago, Chile}

\icmlaffiliation{uv}{Instituto de Física y Astronomía, Universidad of Valparaíso, Valparaíso, Chile}

\icmlaffiliation{cmm}{Center for Mathematical Modelling, Universidad de Chile, Santiago, Chile}

\icmlaffiliation{dai}{Data and Artificial Intelligence Initiative, Universidad de Chile, Santiago, Chile}

\icmlaffiliation{astropuc}{Instituto de Astrofísica, Facultad de Física, Pontificia Universidad Católica de Chile, Santiago, Chile}

\icmlaffiliation{astroing}{Centro de Astroingeniería, Pontificia Universidad Católica de Chile, Santiago, Chile}

\icmlaffiliation{eso}{European Southern Observatory, Garching bei München, Germany}

\icmlaffiliation{die}{Department of Electrical Engineering, Universidad de Chile, Santiago, Chile}

\icmlcorrespondingauthor{Manuel Pérez-Carrasco}{maperezc@inf.udec.cl}

\icmlkeywords{Machine Learning, Anomaly Detection, Astronomy}

\vskip 0.3in
]



\printAffiliationsAndNotice{}  

\begin{abstract}

With the increasing volume of astronomical data generated by modern survey telescopes, automated pipelines and machine learning techniques have become crucial for analyzing and extracting knowledge from these datasets. Anomaly detection, i.e. the task of identifying irregular or unexpected patterns in the data, is a complex challenge in astronomy. In this paper, we propose Multi-Class Deep Support Vector Data Description (MCDSVDD), an extension of the state-of-the-art anomaly detection algorithm One-Class Deep SVDD, specifically designed to handle different inlier categories with distinct data distributions. MCDSVDD uses a neural network to map the data into hyperspheres, where each hypersphere represents a specific inlier category. The distance of each sample from the centers of these hyperspheres determines the anomaly score. We evaluate the effectiveness of MCDSVDD by comparing its performance with several anomaly detection algorithms on a large dataset of astronomical light-curves obtained from the Zwicky Transient Facility. Our results demonstrate the efficacy of MCDSVDD in detecting anomalous sources while leveraging the presence of different inlier categories. The code and the data needed to reproduce our results are publicly available at \url{https://github.com/mperezcarrasco/AnomalyALeRCE}.
\end{abstract}

\section{Introduction}
\label{Intro}

With modern survey telescopes producing unprecedented volumes of data, it has become unfeasible to analyze astronomical data massively by human inspection. Consequently, the necessity for automated pipelines arises, enabling the extraction of knowledge from these vast datasets in a data-driven manner. Among the intriguing and complex astronomical challenges lies the task of anomaly detection (AD), which refers to the identification of irregular or unexpected patterns that deviate from our existing understanding of the data \citep{Chandola_2009}. 

In recent years, the application of machine learning techniques in the field of astronomy has led to remarkable advancements in detecting anomalous sources in a systematic manner. For instance, \citealt{xiong_2010} employed a hierarchical probabilistic model, while \citealt{Baron-Pznanski_2018} utilized an unsupervised Random Forest approach to identify outliers within galaxy spectra from the Sloan Digital Sky Survey (SDSS). More recently, \citealt{villar2020} applied an IForest algorithm to a latent space derived from a Variational Recurrent Autoencoder (VRAE), specifically targeting a simulated dataset of supernovae (SNe). Similarly, \citealt{sanchezsaez_2021} employed a VRAE architecture to analyze the light curves of active galactic nuclei (AGN) within the Zwicky Transient Facility Data Release 5 (ZTF DR5). Furthermore, the work of \citealt{ishida_2021} employed Gaussian Processes to extract features and train Isolation Forest algorithm in an active learning manner using the Open Supernova Catalog dataset and the Photometric LSST Astronomical Time-series Classification Challenge (PLAsTiCC; \citealt{kessler_2019, mlozek}). In a similar vein, \citealt{Muthukrishna_2022} used a bayesian parametric model to find anomalous sources using simulated light curves from SNANA \citep{SNANA_2009}. Previous works share a common training procedure: they utilize a sample of normal and well-identified objects categorized as inliers to train an algorithm that can detect samples deviating from these inliers. This approach, commonly known as one-class anomaly detection \citep{scholkopf_1999}, represents the current state-of-the-art for various anomaly detection problems. However, when different inlier categories with distinct data distributions are present in the training set, it remains unclear how to effectively utilize and leverage these methods.

To address this challenge, we present Multi-Class Deep Support Vector Data Description (MCDSVDD), an extension of the state-of-the-art anomaly detection algorithm One-Class Deep SVDD \citep{ruff_2018}. Our proposed method employs a neural network to map the data into hyperspheres, where each hypersphere encapsulates the data representation of a specific inlier category. Consequently, the "weirdness" of each sample is determined by its distance from the center of the closest hypersphere. To evaluate the effectiveness of our approach, we compare the performance of several one-class anomaly detection algorithms on a large dataset of astronomical light-curves obtained from the Zwicky Transient Facility (ZTF; \citealt{Bellm_2018}). Through our comprehensive evaluation, we demonstrate the efficacy of MCDSVDD in detecting anomalous sources while effectively leveraging the presence of different inlier categories.

\section{Data}
\label{data}

In this work we use data alerts from the from the ZTF data stream \citep{Bellm_2018}. In this data stream, an alert is triggered by an object in the sky whose current (science) image has a significant difference with respect to a template (reference) image \citep{Masci_2018}. For alerts to be streamed by ZTF, they need to pass the cut-off criteria defined by the real/bogus detection system designed by the ZTF Collaboration. These criteria include signal-to-noise ratios, near-edge image positioning, negative and bad pixels, and morphological and photometric features \citep{Mahabal_2019, duev_2019}.

Light-curves are constructed using the same procedure of Section 4.4 of \citet{forster_2020}. We perform crossmatch with the AllWISE\footnote{The AllWISE Data Release can be found at \url{http://wise2.ipac.caltech.edu/docs/release/allwise/}} public source catalog \citep{Wright_2010,Mainzer_2011}, using a matching radius of 2 arcseconds, obtaining W1, W2, and W3 photometry. Then, 152 features are calculated for each light-curve using the ALeRCE broker feature extractor \cite{forster_2020}. These features include detection and non-detection features (see Section 3 of \citealt{sanchezsaez_2020} for details). We only use objects with at least six detections in either $g$ or $r$ bands.

Our training set consisted in labeled light-curves that follow a taxonomy with three main classes (hereafter, the top level classes): transient, stochastic, and periodic. Each of these categories is then subdivided into the following subclasses:

\begin{itemize}
    \item Transient: Type Ia supernova (SNIa), Type Ibc supernova (SNIbc), Type II supernova (SNII), and super-luminous supernova (SLSN);
\item Stochastic: Type 1 Seyfert galaxy (AGN; i.e., host-dominated active galactic nuclei), Type 1 quasar (QSO; i.e., nucleus-dominated AGN), blazar (blazar; i.e, beamed jet dominated AGN), young stellar
object (YSO; including bursters, dippers and purely rotation modulated lightcurves of pre-main sequence stars), and cataclysmic variable/Nova (CV/Nova);
\item Periodic: long-period variable (LPV; includes
regular, semi-regular, and irregular variable stars), RR Lyrae (RRL), Cepheid (CEP), eclipsing binary (E), and $\delta$ Scuti (DSCT).
\end{itemize}

In Section \ref{sec:metholodogy} we explain in detail how the above-mentioned categories are used to train and evaluate our anomaly detection algorithms by hiding the light curves from a given subclass and considering all the others as inliers. This allows for a rigorous comparison of algorithms that helps to select the most promising method to be applied in a real-world scenario.

\section{Method} \label{sec:method}

In this work we present Multi-Class Deep SVDD (MCSVDD), an extension of the One-Class Deep SVDD method \cite{ruff_2018}.

\subsection{One-Class Deep SVDD}

Deep Support Vector Data Description \citep[Deep SVDD;][]{ruff_2018} is a neural network-based approach related to OCSVM \citep{scholkopf_1999}. The general idea is to map  the data from an input space $\mathcal{X} \subseteq \mathbb{R}^{d}$ into a new feature space $\mathcal{F} \subseteq \mathbb{R}^{p}$ using a neural network $\phi(\cdot;\Theta): \mathcal{X} \rightarrow \mathcal{F}$ with $L$ hidden layers and parameters $\Theta = \{\theta^{1},...,\theta^{L}\}$. Specifically, we want the neural network $\phi$ to learn a representation that minimizes the volume of a hypersphere of radius $R>0$ and center $\mathbf{c}$ that enclose the normal data in the output space $\mathcal{F}$.

The objective function for this task is defined by:

\begin{equation}
    \min_{\Theta, R}{ R^{2} + \frac{1}{N} \sum_{i=1}^{N} \max\{0, ||\phi(\mathbf{x}_{i};\Theta) - \mathbf{c}||^{2}} - R^{2}\} + \frac{\lambda}{2} \mathbf{\Theta}^\top\mathbf{\Theta},
\end{equation}
where $\lambda$ is a hyperparameter that controls the weight decay on the network parameters $\theta$. In practice, the encoding part of an autoencoder can be used to map the data into the latent space and define the center $\mathbf{c}$. The parameters $\theta$ and $R$ are alternately and iteratively optimized via gradient descent and line-search respectively.

In cases when training data contains only normal samples, one can simply penalize the squared distance of all the data points with respect to a center $\mathbf{c}$ as follows \cite{ruff_2018, Ruff_2019}:

\begin{equation}
    \min_{\Theta}{\frac{1}{N} \sum_{i=1}^{N} \max\||\phi(\mathbf{x}_{i};\Theta) - \mathbf{c}||^{2}} + \frac{\lambda}{2} \mathbf{\Theta}^\top\mathbf{\Theta}.
\end{equation}

This loss function will be used for comparisons in Section \ref{sec:results}.

\subsection{Multi-Class Deep SVDD}

We extend the One-Class Deep SVDD to take into account the different categories presented in our taxonomy. Instead of modeling a single hypersphere to enclose the normal data, MCSVDD models multiple hyperspheres, where each one corresponds to a given class and enclose normal data for that class. The idea is to learn a neural network that maps objects from the same class close to each other and far from objects from different classes. As abnormal samples come from unseen classes, their distances to each hypersphere should be larger than those of normal datapoints. We named this method Multi-Class Deep SVDD (MCSVDD) as we now have multiple inlier classes.  

Assuming normal data pairs coming from $M$ different classes $y \in \{1,...,M\}$,  the objective function for a training set composed by different inlier classes is given by:

\begin{equation}
    \min_{\mathbf{\Theta}}{ \sum_{j=1}^{M} \frac{1}{N_{j}} \sum_{i=1}^{N} \mathbbm{1}_{(y_{i}=j)}||\phi(\mathbf{x}_{i};\Theta) - \mathbf{c}_{j}||^{2}} + \frac{\lambda}{2}\mathbf{\Theta}^\top\mathbf{{\Theta}},
\end{equation}

where $\lambda$ is a hyperparameter that controls the weight decay regularizer on the network parameters $\mathbf{\theta}$, $\mathbbm{1}_{(y_{i}=j)}$ is an indicator function that becomes $1$ if $y_{i}=j$ and $0$ otherwise, and $N_{j}$ is the number of data points that belong to class $j$.

By following this approach, it is possible to define an anomaly score $A(\cdot)$ based on the distance of the data points to the centers of the hyperspheres as follows:

\begin{equation}
    A(x_{i}) = \min_{j}\{||\phi(\mathbf{x}_{i};\mathbf{\Theta^{*}}) - \mathbf{c}_{j}||^{2}\}.
\end{equation}


where $\Theta^{*}$ are the parameters of the trained neural network. 

\section{Methodology} \label{sec:metholodogy}

\begin{figure}[h]
\centering
\includegraphics[width=.5\textwidth]{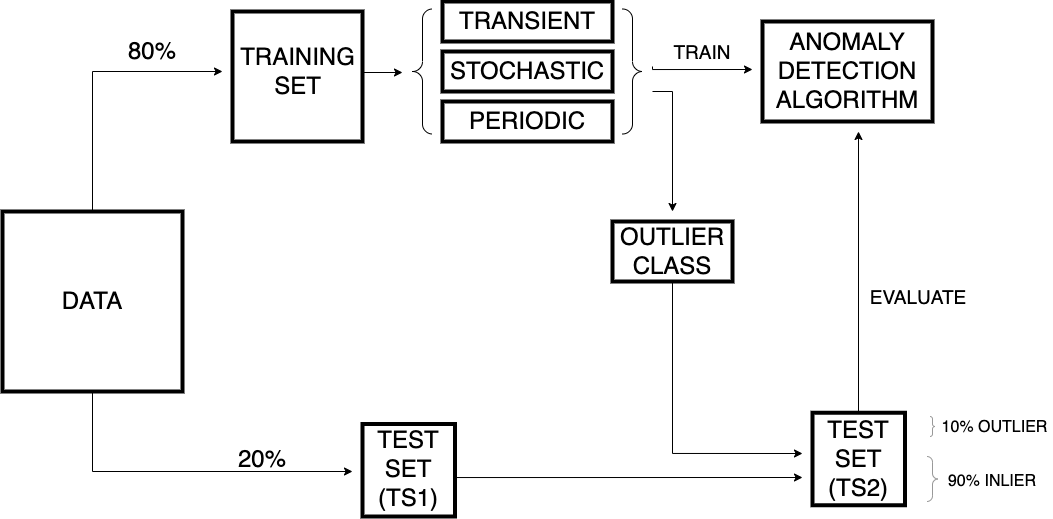} 
\caption{Methodology for training and evaluation of the anomaly detection algorithms. We split the data into a training set and test set, composed by $80\%$ and $20\%$ of the data, respectively. The training set is subdivided into transient, stochastic, and periodic data. For each of these classes, we choose each subclass as the outlier class. The outlier class is removed from the training set and added to the test set (TS2). Then, an anomaly detection algorithm is trained using the remaining objects of each of the classes, and is evaluated using TS2.}
\label{fig:odmethodology}
\end{figure}

For fair comparisons, we follow the same training and evaluation procedure for all our experiments. From the ZTF light-curves, we extracted 152 features as described in Sec. \ref{data}. All the features were normalized between -1 and 1 using the quantile normalization method. We randomly split the data into a training set ($80\%$) and a test set ($20\%$) in a stratified fashion in order to preserve the proportion of samples per class. The training set is divided into five stratified subsets in order to perform 5-fold cross-validation for model selection.
Fig.~\ref{fig:odmethodology} shows a scheme of our training and evaluation methodology.

\begin{table*}[!ht]
  
  \hskip-3.3cm
  \begin{center}
  \scalebox{0.60}{
  \begin{tabular}{|l||cccc||ccccc||ccccc|} 
    \hline 
    & \multicolumn{4}{c||}{ \textbf{Transient}}&
    \multicolumn{5}{c||}{ \textbf{Stochastic}}&
    \multicolumn{5}{c|}{ \textbf{Periodic}} \\

    Method  & SLSN & SNII & SNIa & SNIbc & AGN & Blazar & CV/Nova & QSO & YSO & CEP & DSCT & E & RRL & LPV \\ \hline
    
    IForest  & $0.640$ & $0.721$ & $0.428$ & $0.490$ & $0.573$ & $0.710$ & $0.975$ & $0.468$ & $0.913$ & $0.359$ & $0.295$ & $0.469$ & $0.549$ & $0.971$ \\
    \citep{Liu_2012}
    & $\pm 0.014$ & $\pm 0.021$ & $\pm 0.032$ & $\pm 0.038$ & $\pm 0.017$ & $\pm 0.009$ & $\pm 0.001$ & $\pm 0.016$ & $\pm 0.003$ & $\pm 0.007$ & $\pm 0.012$ & $\pm 0.021$ & $\pm 0.033$ & $\pm 0.007$ \\ \hline
  
    OCSVM  & $0.577$ & $0.587$ & $0.434$ & $0.492$ & $0.532$ & $0.443$ & $0.909$ & $\mathbf{0.517}$ & $0.792$ & $0.432$ & $0.557$ & $0.555$ & $0.539$ & $0.943$ \\ \citep{scholkopf_1999}
    & $\pm 0.014$ & $\pm 0.014$ & $\pm 0.021$ & $\pm 0.011$ & $\pm 0.008$ & $\pm 0.002$ & $\pm 0.001$ & $\mathbf{\pm 0.005}$ & $\pm 0.005$ & $\pm 0.004$ & $\pm 0.005$ & $\pm 0.003$ & $\pm 0.004$ & $\pm 0.001$ \\ \hline
    
    AE  & $\mathbf{0.736}$ & $0.807$ & $0.438$ & $0.537$ & $\mathbf{0.701}$ & $\mathbf{0.762}$ & $\mathbf{0.980}$ & $0.443$ & $\mathbf{0.990}$ & $0.564$ & $0.367$ & $0.864$ & $0.907$ & $\mathbf{0.996}$ \\
    \citep{rumelhart_1987}
     & $\mathbf{\pm 0.022}$ & $\pm 0.021$ & $\pm 0.015$ & $\pm 0.019$ & $\mathbf{\pm 0.010}$ & $\mathbf{\pm 0.006}$ & $\mathbf{\pm 0.016}$ & $\pm 0.004$ & $\mathbf{\pm 0.001}$ & $\pm 0.024$ & $\pm 0.015$ & $\pm 0.009$ & $\pm 0.015$ & $\mathbf{\pm 0.000}$ \\ \hline
    
    VAE  & $0.669$ & $0.690$ & $0.404$ & $0.522$ & $0.596$ & $0.597$ & $0.849$ & $0.500$ & $0.795$ & $0.442$ & $0.417$ & $0.561$ & $0.451$ & $0.936$ \\
    \citep{kingma_2013}
     & $\pm 0.015$ & $\pm 0.023$ & $\pm 0.018$ & $\pm 0.025$ & $\pm 0.007$ & $\pm 0.010$ & $\pm 0.028$ & $\pm 0.009$ & $\pm 0.009$ & $\pm 0.010$ & $\pm 0.007$ & $\pm 0.007$ & $\pm 0.006$ & $\pm 0.007$ \\ \hline
    
    Deep SVDD  & $0.644$ & $0.731$ & $0.475$ & $0.507$ & $0.496$ & $0.607$ & $0.932$ & $0.411$ & $0.901$ & $0.707$ & $0.482$ & $0.636$ & $0.774$ & $0.785$ \\
    \citep{ruff_2018}
    & $\pm 0.043$ & $\pm 0.043$ & $\pm 0.040$ & $\pm 0.040$ & $\pm 0.025$ & $\pm 0.044$ & $\pm 0.015$ & $\pm 0.008$ & $\pm 0.022$ & $\pm 0.027$ & $\pm 0.054$ & $\pm 0.055$ & $\pm 0.068$ & $\pm 0.025$ \\ \hline
    
    MCDSVDD  & $0.686$ & $\mathbf{0.828}$ & $\mathbf{0.624}$ & $\mathbf{0.584}$ & $0.706$ & $0.512$ & $0.770$ & $0.483$ & $0.854$ & $\mathbf{0.858}$ & $\mathbf{0.819}$ & $\mathbf{0.945}$ & $\mathbf{0.953}$ & $0.953$ \\
    (This work)
    & $\pm 0.051$ & $\mathbf{\pm 0.024}$ & $\mathbf{\pm 0.039}$ & $\mathbf{\pm 0.032}$ & $\pm 0.069$ & $\pm 0.113$ & $\pm 0.127$ & $\pm 0.080$ & $\pm 0.041$ & $\mathbf{\pm 0.025}$ & $\mathbf{\pm 0.015}$ & $\mathbf{\pm 0.006}$ & $\mathbf{\pm 0.003}$ & $\pm 0.008$ \\ \hline

  \end{tabular} }
  \end{center}

    \caption{Evaluation of the performance of each model when applied to each of the top level taxonomy (transient, stochastic, periodic). Each row represents a different outlier detection algorithm, and each column represents the subclass considered as outlier. The performance is evaluated using the AUROC scores. Best metrics per class are marked in boldface}
  \label{table:results}
\end{table*}

\subsection{Training}\label{sec:training}

To facilitate the problem at hand, our training dataset is divided into three primary classes: transient, stochastic, and periodic. These classes are further stratified into 14 subclasses, as described in Section \ref{data}. For each main class, we employ a dedicated anomaly detector, resulting in three separate detectors. Since anomalies are typically unknown in real-world scenarios, during cross-validation, we designate a subclass from each main class as the anomalous class and exclude it from the training set. Consequently, the model is trained on the remaining subclasses, with the excluded subclass reserved solely for evaluation purposes. This approach ensures that the model does not incorporate data from the chosen anomalous subclass during the training phase. The process is iterated for each subclass, thereby providing a comprehensive evaluation of the anomaly detectors' overall performance. The assessment primarily focuses on the detectors' ability to successfully identify the removed subclasses, aligning with established practices in the machine learning literature \citep{ruff_2018}.

\subsection{Evaluation}

Although real outlier events are typically unavailable for evaluation, we aim to create a realistic scenario to identify the most promising anomaly detection models. As previously mentioned, 20\% of the dataset is designated as the test set (TS1) and is never utilized for training any model. We create a second test set (TS2), including all the objects from TS1 that belong to the inlier subclasses, and outliers from both TS1 and those who were removed from the training set. To ensure a realistic distribution, TS2 is composed of 10\% outliers and 90\% inliers. This setup enables a comprehensive assessment of the models' ability to detect the chosen outlier class within TS2. See Figure \ref{fig:odmethodology} for an illustrative diagram of training en evaluation methodology.

To evaluate model performance when a subclass is treated as an outlier, we employ the Area Under the Receiver Operating Characteristic curve (AUROC; \citealt{davis_2006}).\\

\section{Results} \label{sec:results}

We conducted a comprehensive comparison of our proposed method, Multi-Class Deep SVDD (MCDSVDD), against five state-of-the-art algorithms for anomaly detection in the field of astronomy. The baseline algorithms we considered were Isolation Forest (IForest; \citealt{Liu_2012}), One-class Support Vector Machine (OCSVM; \citealt{scholkopf_1999}), Autoencoder (AE; \citealt{rumelhart_1987}), Variational Autoencoders (VAE; \citealt{kingma_2013}), and Deep Support Vector Data Description (Deep SVDD; \citealt{ruff_2018}). See Appendix I for details and hyperparameters of all the algorithms used in this work. Table \ref{table:results} presents the 5-fold AUROC values obtained by each anomaly detection model on the TS2 dataset. The table showcases the performance of each model when considering each of the 14 subclasses as an outlier in separate trials. The highest AUROC values for each subclass are highlighted in bold, indicating the best-performing model for that particular task.

As can be seen, our proposed method MCDSVDD consistently outperforms all other methods  we considered for the transient and periodic objects. Our method is able to detect SNII, SNIa, and SNIb as outliers with higher AUROC values than the rest. For the SLSN subclass, AE's performance is statistically indistinguishable from MCDSVDD (i.e., not statistically significant; $p$-value = 0.0789). In terms of the periodic classes, MCDSVDD shows superior performance in detecting CEP, DSCT, E, and RRL as outliers, while AE demonstrates better performance in detecting LPVs. Regarding stochastic sources, AE outperforms the other methods for four out of the five subclasses (AGN, Blazar, CV/Nova, and YSO), while OCSVM achieves the highest performance for QSOs. It is worth mentioning that QSO light curves often exhibit slow and smooth temporal variations, making them more challenging to detect as anomalies. Therefore, AE emerges as the best outlier detection algorithm for stochastic sources, whereas MCDSVDD excels in detecting transient and periodic sources. We hypothesize that the high variability of stochastic light curves may impact MCDSVDD's capability to detect different types of sources, unlike AE, which benefits from its reconstruction loss function.

\section{Conclusion} \label{sec:conclusion}
In this paper, we addressed the challenge of effectively utilizing different inlier categories with distinct data distributions in anomaly detection for astronomical data. We proposed Multi-Class Deep SVDD (MCDSVDD), an extension of the state-of-the-art One-Class Deep SVDD algorithm. MCDSVDD employs a neural network to map the data into hyperspheres, where each hypersphere represents a specific inlier category. By measuring the distance of each sample from the centers of these hyperspheres, MCDSVDD determines the anomaly scores of the samples. We evaluated the performance of MCDSVDD along with several one-class anomaly detection algorithms on a large dataset of astronomical light-curves from the Zwicky Transient Facility (ZTF). The results demonstrated that MCDSVDD outperforms other algorithms in detecting anomalous sources while effectively leveraging the presence of different inlier categories. Our findings highlight the importance of considering distinct inlier categories in anomaly detection tasks and showcase the potential of MCDSVDD for identifying anomalies in astronomical data. Finally, the code and data to reproduce the results presented in this work are publicly available at \url{https://github.com/mperezcarrasco/AnomalyALeRCE}.\\

\section{Acknowledgements}

This work gratefully acknowledges funding from ANID through: 
Millennium Science Initiative Programs ICN12\_009 (MPC, GCV, LHG, FF, AMMA, IR, FEB, MC, PAE, GP) and NCN19\_171 (AB); 
CATA-Basal AFB-170002 (FEB, MC), ACE210002 (FEB, MC) and FB210003 (FEB, MC, FF); 
BASAL project FB210005 (AMMA);
FONDECYT Regular 1231877 (GCV), 1200710 (FF), 1190818 (FEB), 1231637 (MC), and 1200495 (FEB);
FONDECYT Postdoctorado 3200250 (PSS);
and infrastructure funds QUIMAL140003 and QUIMAL190012; FONDECYT 1220829 (PAE).
We acknowledge support from REUNA Chile, which hosts and maintains some of our infrastructure. This work has been
possible thanks the use of AWS-U.Chile-NLHPC credits and computational resources provided by Data Observatory. Powered@NLHPC: This research was partially
supported by the supercomputing infrastructure of the NLHPC (ECM-02).
This research makes use of the community built resource VSX available at the AAVSO. We additionally thank G. Hajdu for useful discussions.


\bibliography{example_paper}
\bibliographystyle{icml2023}

\newpage
\appendix
\onecolumn
\section{Appendix I}
In this work, six anomaly detection algorithms are examined in order to compare their performances in finding
outliers. The algorithms and their respective hyperparameters are:

\textbf{Isolation Forest}

The Isolation Forest algorithm is based on the concept of isolation trees. It randomly selects features and divides them into distinct non-overlapping regions using a randomly selected threshold criterion. The anomaly score is proportional to the number of splits required to isolate each object in the sample. Anomalous objects are expected to require fewer splits to be isolated. To address the issue of overfitting, Isolation Forest employs ensembles of isolation trees.

We utilized the implementation of Isolation Forest provided by scikit-learn. The hyperparameters were set as follows: number of trees = 100, number of samples to train each base estimator = 256, and contamination parameter = 0.1. These values were recommended in the original work by Liu et al. (2012). Since outliers were not used for training, we did not perform hyperparameter selection through cross-validation.

\textbf{One-Class Support Vector Machine}

The One-Class Support Vector Machine (OCSVM) is an anomaly detection method based on Support Vector Machines (SVM). OCSVM maps the data into a new feature space, where the inner product between two objects can be represented with a kernel function, such as the Gaussian kernel. It learns a hyperplane in this feature space that separates the region where most of the data lie. During testing, the anomaly score is computed by evaluating the distance of data points with respect to the learned hyperplane.

We used the One-Class SVM implementation provided by scikit-learn with the radial basis function (RBF) kernel. The hyperparameters were set to $\nu = 0.01$ and the contamination parameter $c = 0.1$, following the default values. As the anomalous samples were assumed to be unknown, we did not perform hyperparameter selection and used the default settings.

\textbf{Autoencoder}

Autoencoders (AE) are unsupervised neural network algorithms that aim to reconstruct the input data using a lower-dimensional representation called the latent space. They consist of an encoder function that maps the input data to the lower-dimensional representation and a decoder function that reconstructs the original data from the latent space. AE's effectiveness in anomaly detection is attributed to the assumption that anomalies are incompressible and cannot be effectively reconstructed from low-dimensional projections. The reconstruction error, computed as the mean squared error, serves as the anomaly score.

We implemented the Autoencoder using PyTorch 1.0.0. Our model architecture consisted in an encoder with 4 hidden layers with \{512, 256, 128, 64\} neurons, and a decoder with \{64, 128, 256, 512\} neurons. We use batch normalization and Leaky ReLU as activation function, except for the last layer that used a Tanh activation function. We selected the hyperparameters by considering the reconstruction error over a validation set consisting only of inliers.

\textbf{Variational Autoencoder}

Variational Autoencoders (VAE) extend autoencoders by incorporating a regularization term. This term enforces the latent space to follow a known distribution, typically a normal distribution. By generating multiple reconstructions for each sample and averaging their reconstruction errors, VAEs compute an anomaly score.

We implemented the Variational Autoencoder using PyTorch 1.0.0. We use the same model architecture as the autoencoder, but we add two extra layers at the end of the encoder to learn the parameters $\mu$ and $\sigma$ of the normal distribution. We selected the hyperparameters using the unsupervised loss function over a validation set consisting only of inliers.

\textbf{Deep Support Vector Data Description}

To train the Deep Support Vector Data Description (Deep SVDD), an Autoencoder is trained until convergence. Then, the decoder is removed, and the center of the hypersphere is estimated as the average of the encoder's outputs on the training data. Subsequently, the parameters of the encoder are re-optimized using the pretrained parameters, with the objective of minimizing the distance between the encoder's output and the center of the hypersphere, along with a weight decay regularizer.

During testing, the anomaly score is defined as the distance between each data point and the center of the hypersphere.

We implemented Deep SVDD using PyTorch 1.0.0. We use the same model architecture as the encoder of the autoencoder. The hyperparameter $\lambda = 0.5 \times 10^{-6}$ was selected by measuring the unsupervised loss function over a validation set of inliers.

\textbf{Multi-Class Deep SVDD}

Similar to Deep SVDD, an Autoencoder is trained until convergence, and the decoder is removed. The centers of the hyperspheres are estimated as the averages of the encoder's outputs, considering only the data points of each class. The parameters of the encoder are re-optimized using a similar objective as Deep SVDD, but considering multiple hyperspheres.

During testing, the anomaly score is determined by measuring the distance between each data point and the center of its closest hypersphere.

We implemented MCDSVDD using PyTorch 1.0.0. We use the same model architecture as the encoder of the autoencoder.  We set the hyperparameter $\lambda = 0.5 \times 10^{-6}$ through cross-validation of the unsupervised loss function over a validation set consisting only of inliers.


\end{document}